\documentclass{ecai} 

\usepackage{booktabs}
\usepackage{latexsym}
\usepackage{amssymb}
\usepackage{amsmath}
\usepackage{amsthm}
\usepackage{booktabs}
\usepackage{enumitem}
\usepackage{graphicx}
\usepackage{color}

\newcommand{\BibTeX}{B\kern-.05em{\sc i\kern-.025em b}\kern-.08em\TeX}

\begin{document}


\begin{frontmatter}


\paperid{123} 


\title{AgentSGEN: Multi-\textbf{Agent} LLM in the Loop for \textbf{S}emantic Collaboration and \textbf{GEN}eration of Synthetic Data}




\author[A]{\fnms{Vu Dinh}~\snm{Xuan}\footnote{Equal contribution.}}
\author[A]{\fnms{Hao}~\snm{Vo}\footnotemark}
\author[B]{\fnms{David}~\snm{Murphy}}
\author[B]{\fnms{Hoang D.}~\snm{Nguyen}\thanks{Corresponding author. Email: hn@cs.ucc.ie.}}

\address[A]{University of Information Technology, VNU–HCM, Ho Chi Minh City, Vietnam}
\address[B]{University College Cork, Cork, Ireland}


\begin{abstract}
The scarcity of data depicting dangerous situations presents a major obstacle to training AI systems for safety-critical applications, such as construction safety, where ethical and logistical barriers hinder real-world data collection. This creates an urgent need for an end-to-end framework to generate synthetic data that can bridge this gap. While existing methods can produce synthetic scenes, they often lack the semantic depth required for scene simulations, limiting their effectiveness. To address this, we propose a novel multi-agent framework that employs an iterative, in-the-loop collaboration between two agents: an Evaluator Agent, acting as an LLM-based judge to enforce semantic consistency and safety-specific constraints, and an Editor Agent, which generates and refines scenes based on this guidance. Powered by LLM's capabilities to reasoning and common-sense knowledge, this collaborative design produces synthetic images tailored to safety-critical scenarios. Our experiments suggest this design can generate useful scenes based on realistic specifications that address the shortcomings of prior approaches, balancing safety requirements with visual semantics. This iterative process holds promise for delivering robust, aesthetically sound simulations, offering a potential solution to the data scarcity challenge in multimedia safety applications.


\end{abstract}

\end{frontmatter}


\section{Introduction}

Safety in workplaces is a critical issue, especially when dealing with rare and extreme indoor scenarios such as blocked emergency exits. Such safety-critical cases are inherently challenging to capture in real-world datasets due to ethical and practical constraints \cite{ding2023survey, neuhausen2020synthetic}. As a result, synthetic data generation has emerged as a promising approach to simulate these dangerous situations in a controlled and repeatable manner, thereby supporting the development and evaluation of safety-critical AI systems \cite{lee2023game, kim2024image}.

\begin{figure*}[t]
  \centering
  \includegraphics[width=0.9\linewidth]{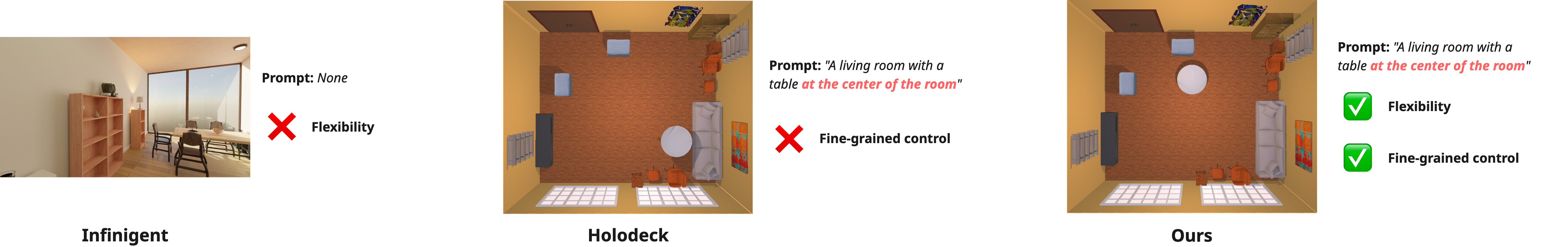}
  \caption{Comparison of 3D scene generation methods.}
  \label{fig:intro_comparison}
  \vspace{1cm}
\end{figure*}

As shown in Figure \ref{fig:intro_comparison}, traditional methods for generating 3D scenes often rely on manual design or procedural algorithms. Manual creation is time-consuming and labor-intensive, while procedural techniques exemplified by systems such as Infinigen \cite{raistrick2024infinigen} offer limited flexibility due to their fixed, algorithm-based solutions that are hard to modify for nuanced safety requirements. More recent methods, such as Holodeck \cite{yang2024holodeck}, leverage large language models to convert high-level natural language prompts into photorealistic 3D scenes. Although Holodeck delivers visually compelling outputs with minimal user input, it lacks the fine-grained control necessary for precisely simulating targeted safety conditions.

%


To address these shortcomings, we propose a multi-agent framework that augments the high-fidelity scene generation process with an iterative semantic guidance process guided by two LLM agents: an Evaluator and an Editor. The Evaluator agent interprets user-defined safety objectives and strategically plans scene modifications by reasoning over an object-centric scene graph. These high-level edit plans are then executed by the Editor agent, which performs concrete scene alterations such as repositioning objects or introducing hazards. This separation of planning and execution allows for fine-grained, goal-driven scene editing while maintaining visual realism and semantic consistency. By iterating through this Actor–Evaluator loop, the system progressively transforms initial scenes into safety-critical environments aligned with user specifications.

For instance, to simulate a hazardous scenario such as a blocked fire exit, the Evaluator agent first interprets the user-defined objectives e.g., obstructing the primary evacuation route with realistic objects by analyzing spatial layouts, object affordances, and proximity constraints within the scene graph. It formulates a detailed action plan specifying which objects to reposition or introduce, ensuring that the obstruction is both semantically and physically coherent. The Editor agent then executes this plan by performing targeted scene modifications, such as moving furniture or placing additional items near the exit, while adhering to constraints like object scale, collision avoidance, and contextual realism. This iterative interaction continues until the scene accurately reflects the desired safety violation. Throughout the process, visual feedback mechanisms such as 2D layout maps and object trajectory visualizations aid in verifying that key conditions like complete blockage of the main exit and preserved accessibility of secondary exits are satisfied. By decoupling strategic reasoning from low-level scene editing, our framework enables precise, context-aware simulation of safety-critical scenarios that are difficult to replicate using conventional AI-driven scene generation methods.

In summary, this paper presents the following contributions:
\begin{itemize}
    \item We develop an end-to-end Actor–Evaluator multi-agent system for controllable 3D scene editing in safety-critical indoor scenarios.
    \item We introduce semantic constraints to maintain realistic spatial relationships and safety rules, ensuring each scene is both visually and semantically aligned.
    \item We provide a diverse dataset of synthetic 3D scenes with blocked fire escapes for AI-based construction safety applications.
    \item Comprehensive evaluations with human raters on randomized images with insights on human and human/AI agreement, demonstrating superior performance in task completion and effectiveness.
\end{itemize}


\section{Related Work}
\subsection{Synthetic data generation}

In safety-critical AI systems, synthetic data generation has become indispensable due to the scarcity of real-world examples involving rare or hazardous events, such as blocked fire exits, equipment failures, or collisions. Collecting such data ethically and practically is often infeasible, especially in domains like autonomous driving or construction safety \cite{lee2023game, kim2024image}. This challenge is particularly acute in computer vision, where models rely on diverse, high-quality visual inputs to detect and respond to complex scenes. Synthetic data offers a controllable and scalable solution, enabling the creation of edge-case scenarios that are otherwise difficult to capture, thereby improving the reliability and generalization of vision-based systems \cite{song2023synthetic, shorten2019survey}.

Traditional data augmentation techniques have been employed to artificially expand training datasets and enhance model generalization. These manual methods include geometric transformations such as flipping, cropping, rotation, and scaling, as well as color space adjustments to modify brightness, contrast, and saturation. Advanced strategies like MixUp \cite{zhang2017mixup} and Mosaic \cite{bochkovskiy2020yolov4} further combine multiple images to create novel training samples, thereby introducing greater variability into the dataset . Despite their utility, these approaches have limitations; they often require significant manual effort to design and may not sufficiently capture the complexity of real-world data distributions, particularly in safety-critical applications where rare events are underrepresented. 

Recent advances in generative modeling techniques have opened promising avenues for the creation of synthetic datasets using state-of-the-art image-generation models, including GAN-based approaches \cite{esser2021taming}, Diffusion-based methods \cite{ zhang2023sine, kawar2023imagic, rombach2022high, podell2023sdxl, sauer2024adversarial,blattmann2023stable}, and Transformer-based frameworks \cite{esser2024scaling} that can produce visually compelling content from minimal prompts; however, their intrinsic operation at the image level often limits fine-grained control and structured label generation, which are critical in safety-critical applications. To address these constraints, extensions such as T2IAdapter \cite{ mou2024t2i} and ControlNet \cite{zhang2023adding} have been introduced to incorporate conditional guidance into the generation process, enabling enhanced control over both the visual content and the accompanying semantic annotations. Despite their potential, the integration of these techniques necessitates a complex multi-stage pipeline that combines generation, conditional adjustment, and post-processing to reliably simulate rare or hazardous events found in domains such as autonomous driving and construction safety, ultimately contributing to the robustness and efficacy of vision-based AI systems.

To enhance control, some methods turn to 3D scene generation, where entire environments are constructed and rendered. Infinigen \cite{raistrick2024infinigen} uses procedural algorithms to synthesize diverse, photorealistic indoor scenes, but its rule-based design limits customization for safety-specific cases. Holodeck \cite{yang2024holodeck} introduces LLM-guided 3D scene synthesis via an anchor-object strategy, reducing manual input. Yet, it lacks flexibility and fine-grained editability of anchor specification, with users unable to incrementally adjust scenes or enforce complex relational constraints.

Our work addresses these gaps by introducing a multi-agent LLM-driven framework that enables iterative, goal-driven scene editing. Unlike prior work, we support fine-grained spatial modifications using lightweight 2D projections, making our approach resource-efficient and broadly applicable. Moreover, by leveraging general-purpose LLMs, we avoid reliance on domain-specific rules or specialized models, offering a scalable and extensible solution for generating safety-critical synthetic data.

\subsection{Multi-Agent Systems}

Recent advances in large language models (LLMs) have significantly expanded the capabilities of artificial intelligence, allowing systems to perform complex tasks such as reasoning, planning, and decision-making, often reaching levels arguably comparable to human performance \cite{tran2025multi, wang2024survey}. However, when these tasks require coordination, contextual awareness, or specialized modularity, single-agent systems often face limitations in scalability, memory, and depth of reasoning \cite{hagendorff2023human}. To overcome these challenges, there is growing interest in multiagent systems (MAS), where multiple LLM-based agents, each with distinct roles, memory, goals, and tools, work together to solve problems that surpass the capabilities of any single model. This collaborative structure supports a more deliberate and reflective form of computation, often referred to as 'slow thinking', which mirrors aspects of human cognitive processes and enables deeper reasoning through iterative interaction.

LLM-based MASs are capable of supporting diverse collaboration modes - including cooperation, competition, and coopetition - each suited to different problem settings \cite{tran2025multi}. For example, MetaGPT \cite{hong2023metagpt} utilize predefined agent roles and Standard Operating Procedures (SOPs) to facilitate structured task decomposition and reduce cascading errors. Conversely, more dynamic systems like DyLAN \cite{liu2023dynamic} emphasize adaptability by selecting agents and assigning roles at runtime based on performance feedback. These architectures not only improve robustness and interpretability but also scale effectively to real-world scenarios across domains such as software development, planning, and safety-oriented scene generation.

In this work, we adopt a multi-agent architecture to address the key limitations of monolithic scene generation approaches - particularly their lack of semantic controllability and limited alignment with safety objectives. By embedding reasoning and constraint-enforcement capabilities within a dedicated Evaluator agent and assigning scene modification responsibilities to an Editor agent, our framework mirrors human-like collaborative workflows. This design enables precise, constraint-aware synthetic data generation, advancing the development of reliable and interpretable AI systems for high-stakes applications.


\section{Method}





This section details the architecture and methodology of \textbf{AgentSGEN}, a multi-agent system designed for generating semantically compliant synthetic data in structured 3D scenes. 
The proposed method is grounded in the cognitive framework of \textit{Dual Process Theory} (DPT) \cite{bellini2022dpt}, which distinguishes between two complementary types of cognitive processing: a slow, deliberative, and reasoning-based System~2, and a fast, intuitive, and reactive System~1. Inspired by this theory, AgentSGEN operationalizes these modes through two distinct agents:

\begin{itemize}
  \item \textbf{Evaluator Agent (System 2)}: A reasoning-intensive agent that formulates high-level plans and validates semantic and safety constraints.
  \item \textbf{Editor Agent (System 1)}: A reactive, low-latency executor responsible for scene manipulation and fine-grained object placement.
\end{itemize}

The system workflow is divided into three major stages: (i) \textit{Context Enrichment}, (ii) \textit{Semantic Planning}, and (iii) \textit{Interactive Scene Editing}. These stages interact with a centralized Scene Graph and Scene Graph Renderer (SGRender) that provide symbolic and visual representations of the environment.

\begin{figure}[t]
  \centering
  \includegraphics[width=\linewidth]{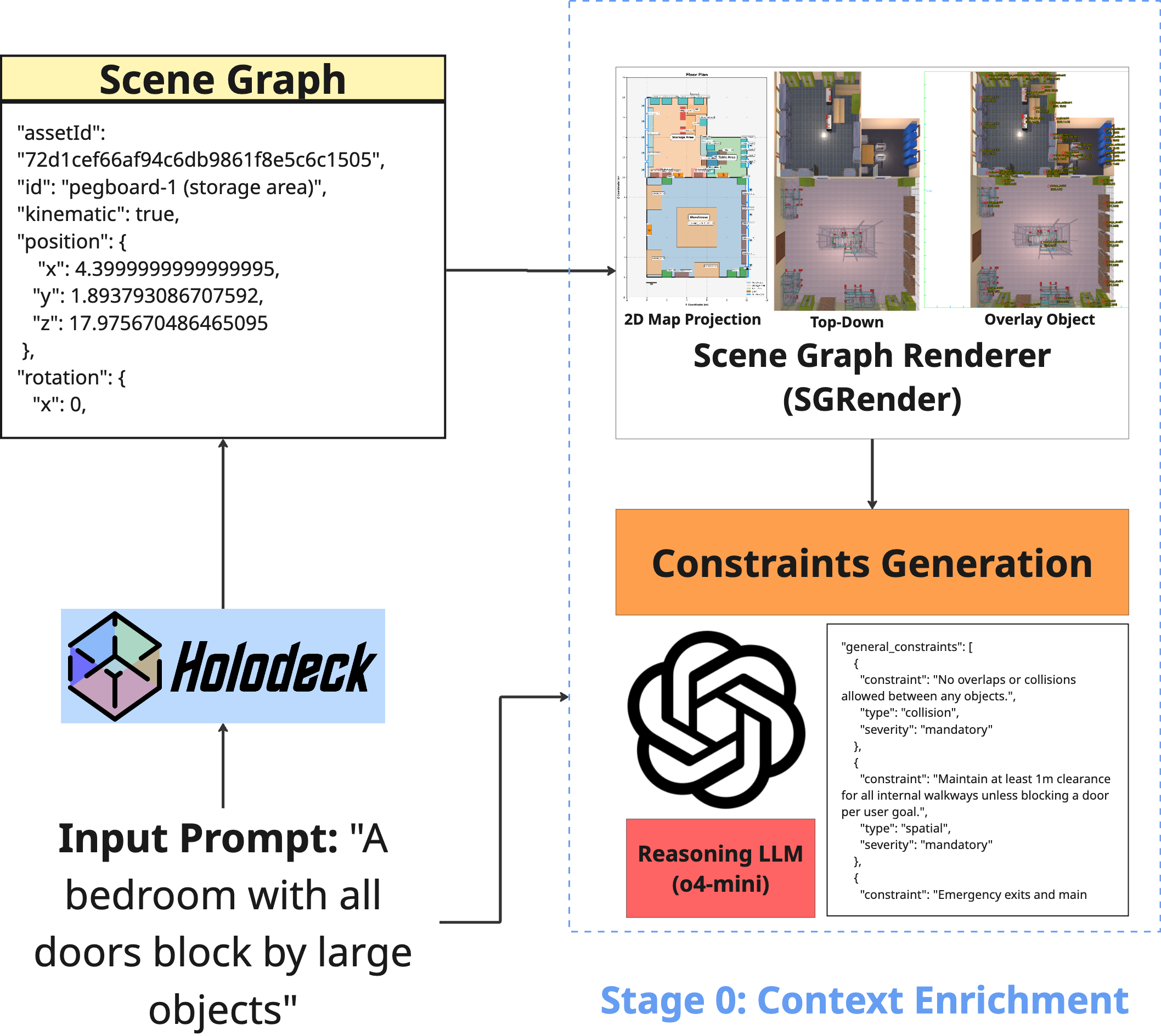}
  \caption{Initial scene from Holodeck with corresponding scene graph and SGRender views.}
  \label{fig:stage0}
  \vspace{1cm}
\end{figure}

\subsection{Problem Formulation}

Given a high-level natural language goal prompt $G$, such as \textit{``a bedroom with doors blocked by large objects''}, the objective is to generate a structured scene $S$ that:

\begin{enumerate}
  \item Aligns semantically with $G$;
  \item Satisfies domain-specific safety and spatial constraints $\mathcal{C}$;
  \item Meets visual and structural coherence;
  \item Supports downstream rendering for synthetic dataset generation.
\end{enumerate}

The process begins with an initial scene graph $\mathcal{G}_0$, which encodes object placements, types, spatial positions, orientations, and physical attributes. Through a sequence of iterative actions $A_1, A_2, \dots, A_T$, the scene is progressively updated to form a final graph $\mathcal{G}_T$, which is subsequently rendered into annotated data formats.

\subsection{Stage 0: Context Enrichment and Constraint Generation}

The initial context is enriched by generating a structured constraint specification $\mathcal{C}$ derived from the user goal $G$ and the initial scene graph $\mathcal{G}_0$. This process is handled by a lightweight LLM (e.g., GPT-4-mini or o4-mini), which outputs a machine-interpretable set of semantic constraints covering four key domains:

\begin{itemize}
  \item \textbf{Collision Constraints}: No overlaps or penetrations between objects.
  \item \textbf{Spatial Constraints}: Minimum clearance for navigable areas, grid-alignment rules, object proximity.
  \item \textbf{Safety Constraints}: Unobstructed access to emergency exits and structural boundaries.
  \item \textbf{Goal Constraints}: Requirements derived directly from the semantics of $G$, e.g., blocking doors with large objects.
\end{itemize}

The output of this phase is a multi-modal context bundle $\mathcal{M} = (\mathcal{G}_0, \mathcal{V}_0, \mathcal{C})$, where $\mathcal{V}_0$ represents visual projections from SGRender, including top-down, 2D projections, and overlay object renderings. This bundle serves as input to both agents in the downstream pipeline.

\begin{figure}[t]
    \centering
    \includegraphics[width=\linewidth]{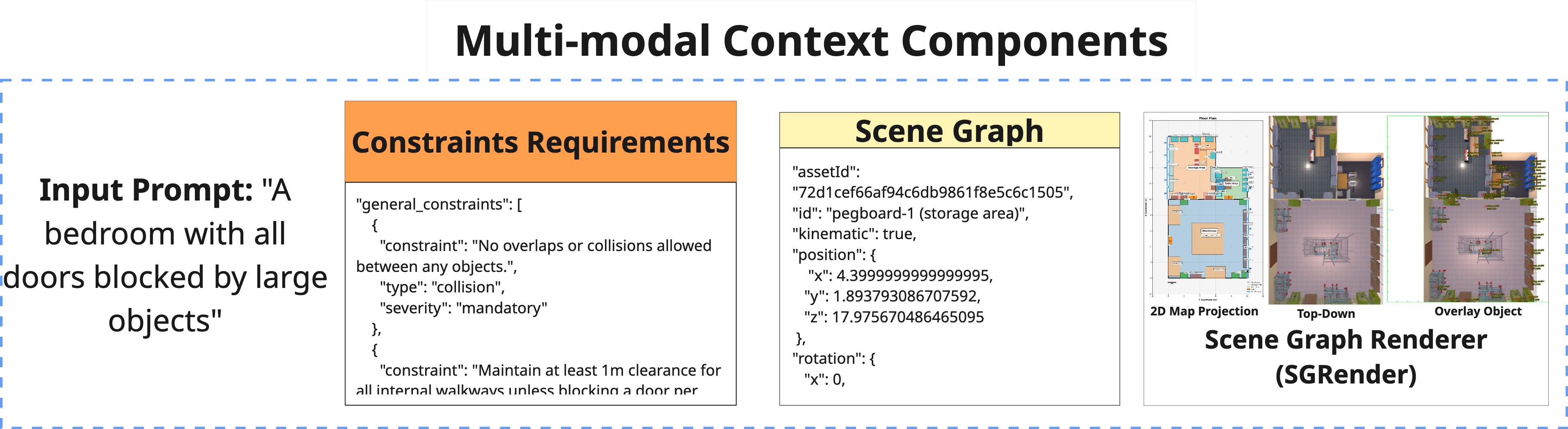}
    \caption{Multi-modal context components passed to both agents. The context bundle consists of (i) goal-specific constraint requirements, (ii) symbolic scene graph, and (iii) 2D/3D renderings from SGRender.}
    \label{fig:context_component}
    \vspace{1cm}
\end{figure}

\subsection{Stage 1: Semantic Planning via Evaluator Agent (System 2)}

The Evaluator Agent is modeled as a \textit{System 2} cognitive entity, capable of slow, deliberate reasoning over symbolic and visual context. It is implemented using a high-capacity reasoning model such as \textbf{O1, O3, or O4-mini with Chain-of-Thought prompting}, and conditioned with a long-form, domain-specific planning prompt.

The Evaluator ingests $\mathcal{M}$ and produces a coherent \textit{Action Plan} $\mathcal{P} = [A_1, A_2, \dots, A_T]$, where each $A_t$ consists of:

\[
A_t = (\text{object\_id}, \text{operation}, \text{parameters})
\]

Each action defines spatial manipulations such as translation, rotation, or deletion of scene elements. The Evaluator may also invoke \textit{internal mental simulation}, including trajectory planning and collision forecasting, leveraging internal visualizations to evaluate the downstream effect of hypothetical actions.

This planning process involves constraint satisfaction checking, rule-based validation, and optimization for semantic alignment. The Evaluator ensures global consistency, feasibility, and alignment with $\mathcal{C}$. Upon completion, the generated action plan is handed over to the Editor Agent for execution.

\subsection{Stage 2: Interactive Scene Editing via Editor Agent (System 1)}

The Editor Agent embodies a \textit{System 1} architecture designed for fast, reactive decision-making. It is implemented using non-reasoning LLMs such as \textbf{GPT-4o or GPT-4.1}, which are well-suited for low-latency execution of atomic actions in constrained environments.

At each timestep $t$, the Editor receives the current state $(\mathcal{G}_t, \mathcal{V}_t)$ and executes a single action $A_t$ from the Evaluator's plan. After execution, the Scene Graph is updated to $\mathcal{G}_{t+1}$, and SGRender produces a new visualization $\mathcal{V}_{t+1}$ that is passed back into the agent.

\begin{figure*}[t]
\centering
    \includegraphics[width=0.9\linewidth]{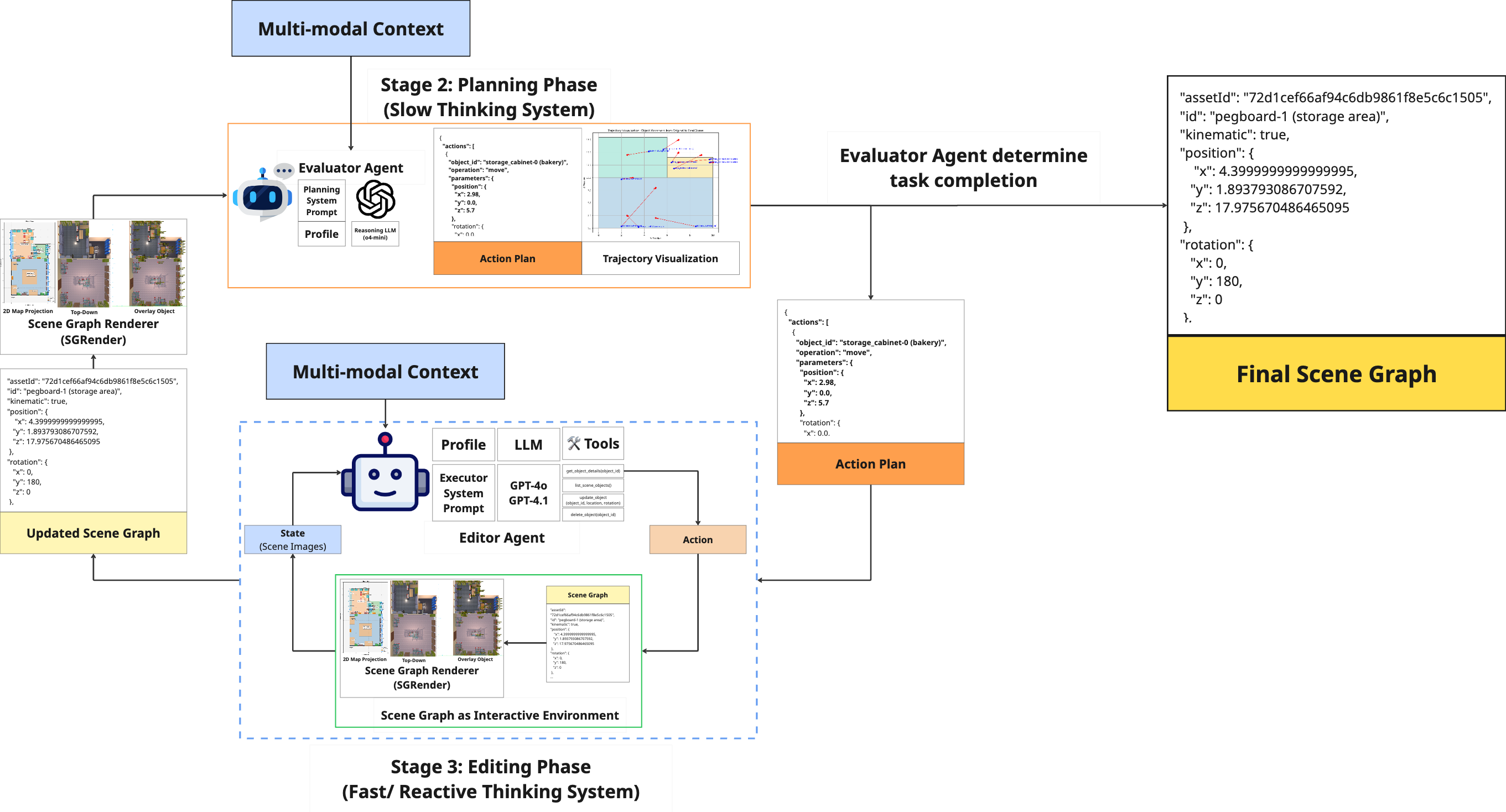}
    \caption{AgentSGEN architecture showcasing semantic planning by the Evaluator Agent and iterative scene editing by the Editor Agent.}
\label{fig:stage_optimize}
\vspace{1cm}
\end{figure*}


The Editor operates with a minimal, task-specific prompt that avoids high-level reasoning and focuses instead on precise object control. As with the Evaluator, the Editor is restricted to \textbf{move}, \textbf{rotate}, and \textbf{delete} operations, ensuring compliance with constraint integrity.

The separation between System 1 and System 2 roles allows the overall framework to benefit from both \textit{semantic depth} and \textit{execution efficiency}. Moreover, this division supports modularity and scalability for future agent extensions or domain-specific adaptations.

\subsection{Scene Graph as Interactive Environment}
The Scene Graph serves not only as a symbolic representation but also as an interactive simulation environment. Each action proposed by the Editor Agent is verified through a lightweight Axis-Aligned Bounding Box (AABB) collision checking mechanism \cite{Ericson2004RealtimeCollision}. The environment supports toggling between collision-aware and collision-disabled modes. This allows for controlled violations of physical realism when justified by high-level goals (e.g., blocking a door intentionally). Feedback is provided to the agent on whether an action results in a valid or invalid move, supporting reactive correction.

This simulation-driven approach bridges symbolic planning with grounded physical validation, enabling a tighter integration between language-based reasoning and geometric feasibility.

\subsection{Limitations of Holodeck and Motivation}
While Holodeck provides a strong foundation for synthetic scene generation, it suffers from several critical limitations (Figure \ref{fig:holodeck_limitation}) that motivate AgentSGEN’s design:

\begin{figure}[t]
  \centering
  \includegraphics[width=\linewidth]{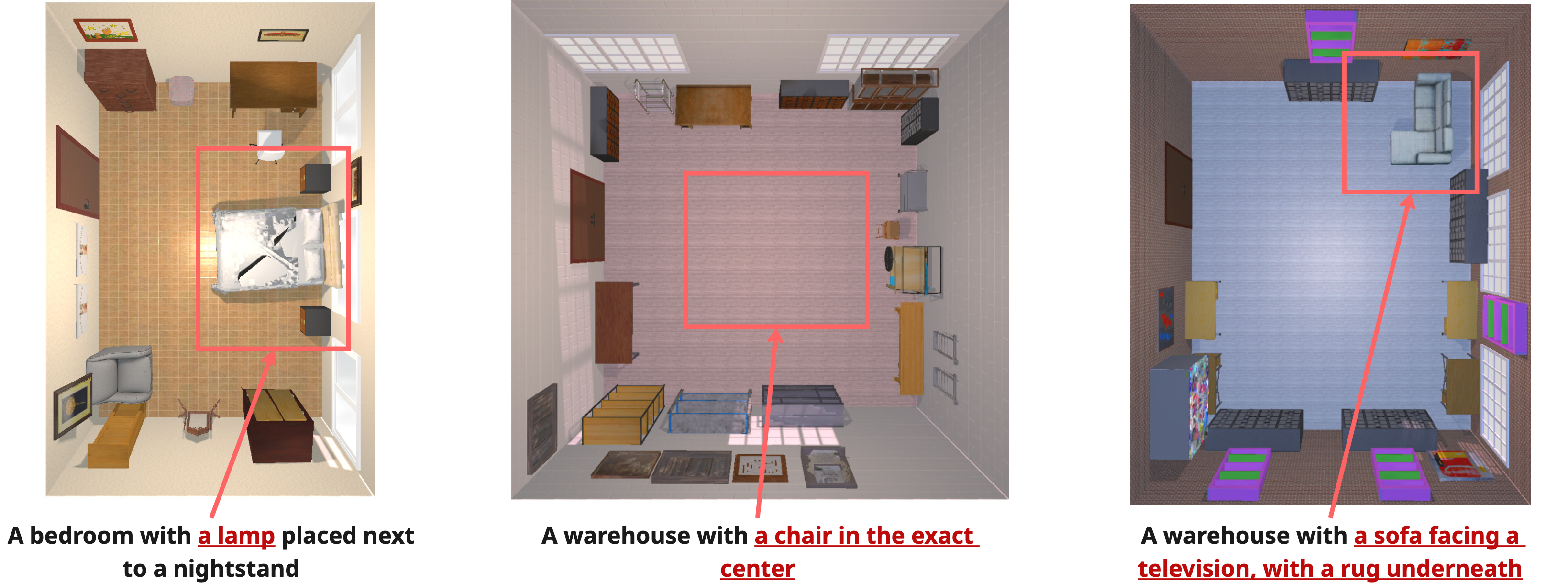}
  \caption{Examples of Holodeck's Limitation}
  \label{fig:holodeck_limitation}
  \vspace{1cm}
\end{figure}

\begin{itemize}
  \item \textbf{Lack of fine-grained control}: Holodeck focuses on procedural scene generation rather than iterative, constraint-based scene editing.

  \item \textbf{Weak semantic alignment}: Scenes often lack adherence to user intent or domain-specific constraints (e.g., safety rules).

  \item \textbf{Non-interactive pipeline}: Once a scene is generated, there is minimal opportunity to refine or reason over object placement.

\end{itemize}

AgentSGEN addresses these gaps by enabling agent-in-the-loop editing, tight semantic constraint enforcement, and modular feedback for correction. Our framework transforms static scene generation into a cognitive reasoning and editing loop.

\subsection{Dual-Agent Cognitive Design}

The cognitive structure of AgentSGEN is explicitly inspired by \textit{Dual Process Theory (DPT)} \cite{bellini2022dpt}. The Evaluator Agent, acting as a deliberate and rule-sensitive planner, performs high-level cognitive tasks including semantic evaluation, constraint checking, and hypothetical reasoning. In contrast, the Editor Agent is a reactive controller that focuses on action execution under the constraints of the current scene state.

This separation aligns with established cognitive architectures where planning and execution are decoupled to enhance robustness, clarity, and adaptability. The Evaluator benefits from longer context windows, hierarchical prompts, and reasoning tokens, while the Editor is optimized for low-latency interactions, loop speed, and fast convergence.

\subsection{Termination, Constraint Validation, and Feedback Loop}

After the final action $A_T$ is executed, the Evaluator Agent re-enters the loop to validate that all constraints $\mathcal{C}$ have been satisfied in the final scene $\mathcal{G}_T$. If violations are detected, the system invokes a feedback loop, updating either the action plan or scene context depending on the source of error.

The feedback mechanism allows AgentSGEN to recover from sub-optimal execution and to iteratively refine the scene until all semantic and safety requirements are met. This loop can be bounded by a maximum iteration threshold or terminated early upon convergence.

\subsection{Final Rendering and Synthetic Dataset Generation}

Once validation is complete, the finalized scene graph $\mathcal{G}_T$ is passed to a high-fidelity render engine (e.g., \textbf{AI2-THOR}, Unity) to produce a complete set of annotated outputs, including:

\begin{itemize}
  \item \textbf{RGB images} from multiple camera angles;
  \item \textbf{Instance segmentation masks} and \textbf{semantic maps};
  \item \textbf{Depth images};
  \item \textbf{Object-level metadata} (bounding boxes, class labels, positions, rotations).
\end{itemize}

\begin{figure*}[ht!]
\centering
\includegraphics[width=0.9\linewidth]{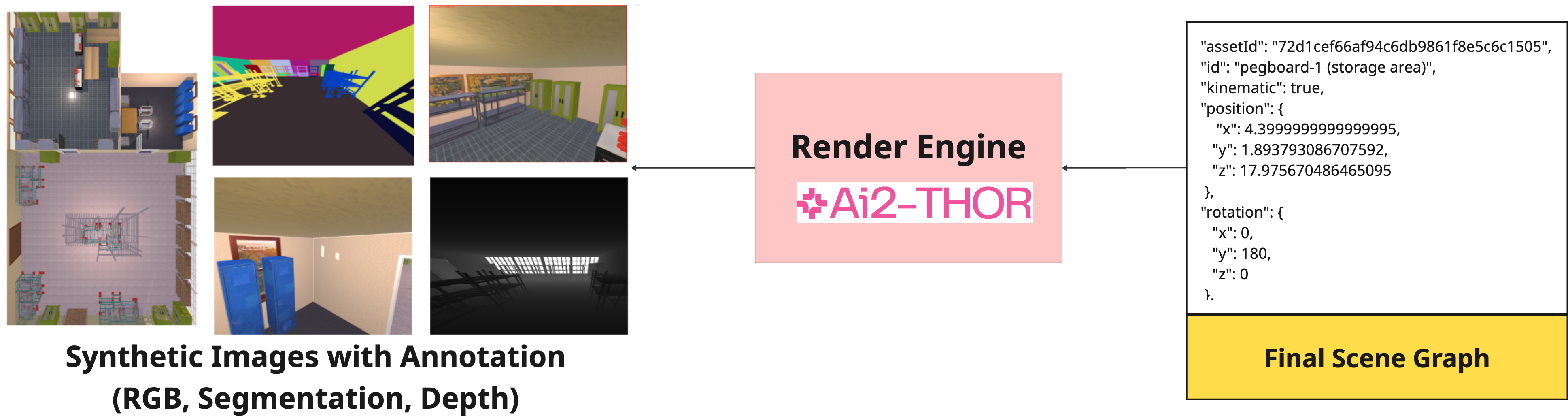}
\caption{Final rendering and synthetic dataset generation using AI2-THOR. Output includes RGB, depth, segmentation, and annotated metadata.}
\label{fig:sdg_stage}
\vspace{1cm}
\end{figure*}

These outputs form a synthetic dataset aligned with the user-defined scenario and ready for training or evaluation of downstream computer vision or robotics models. The dataset is fully reproducible, extensible, and customizable, enabling large-scale generation across diverse safety-critical domains.

In sum, AgentSGEN provides a cognitively grounded, modular, and scalable solution to the challenge of semantically controlled synthetic data generation for safety-critical applications.




\section{Evaluation}

We evaluated our framework across 53 diverse indoor scene types sourced from the MIT Indoor Scenes dataset \cite{mit_53_scenes_quattoni2009recognizing}, encompassing categories such as bedrooms, kitchens, offices, and more. For each room type, we generated an initial scene using the Holodeck procedural generator~\cite{yang2024holodeck}, which ensured standardized and reproducible layouts.

To meet the following safety-critical design goal:

\begin{quote}
\textbf{Goal:} \textit{``A \{room\_type\}, where doors are blocked with large objects.''}
\end{quote}

We applied our editing pipeline under two operational configurations:

\begin{itemize}
    \item \textbf{With Collision Checking:} Enforces physical feasibility and realism by preventing object interpenetration.
    \item \textbf{Without Collision Checking:} Allows free-form object placement to test semantic control under looser constraints.
\end{itemize}

Both versions were compared against the unedited Holodeck baseline to assess how effectively our approach satisfies the task-specific and perceptual criteria. All scenes were rendered and evaluated through structured human annotations.

\subsection{Evaluation Protocol and Annotator Setup}

Two evaluation tasks with randomization were designed to isolate different aspects of performance: (1) binary task completion and (2) goal-oriented semantic quality. Each annotator was trained accordingly:
\begin{enumerate}
    \item \textbf{Binary Task Completion:} Annotators viewed side-by-side renderings (Fig.~\ref{fig:ab_test}) of Holodeck and our edited scene, then selected the one better aligned with the goal. The results were aggregated into a preference matrix (Fig.~\ref{fig:task_completion}).
    
\begin{figure}
    \centering
    \includegraphics[width=\linewidth]{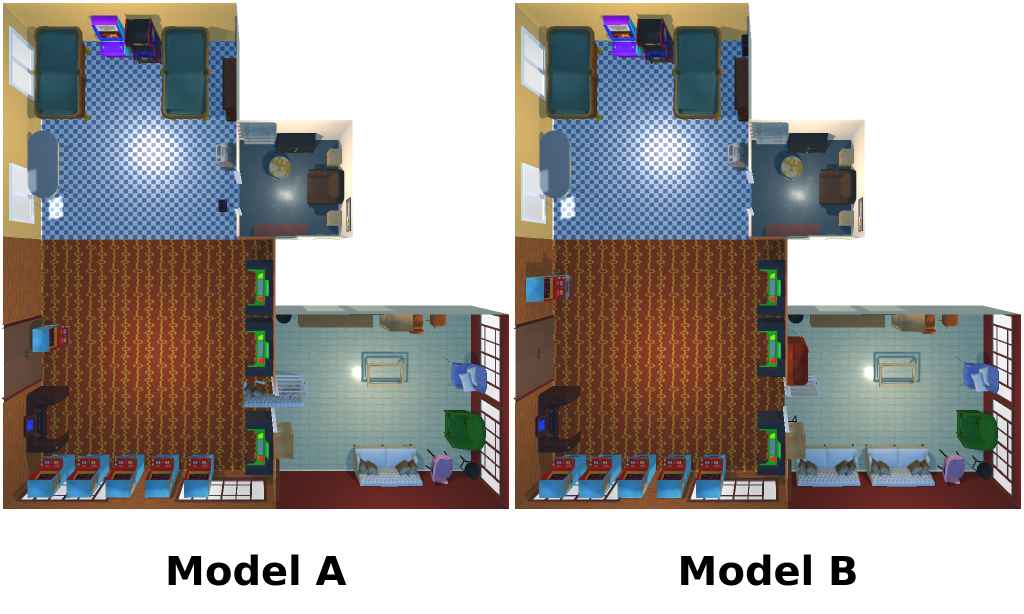}
    \caption{Binary Task Completion for both human evaluation and automatic evaluation setup}
    \label{fig:ab_test}
    \vspace{1cm}
\end{figure}

    \item \textbf{Goal-Oriented Evaluation:} Annotators were shown the prompt in a blind, random order and asked to answer three Likert-scale questions measuring how well the scene satisfies the blocking requirement.

\end{enumerate}

\subsection{Task Completion Results}

To validate consistency, inter-annotator agreement was computed using Cohen’s kappa statistic. A score of 0.406 indicated moderate agreement, supporting the reliability of subjective ratings.

\begin{figure}[t]
    \centering
    \includegraphics[width=0.7\linewidth]{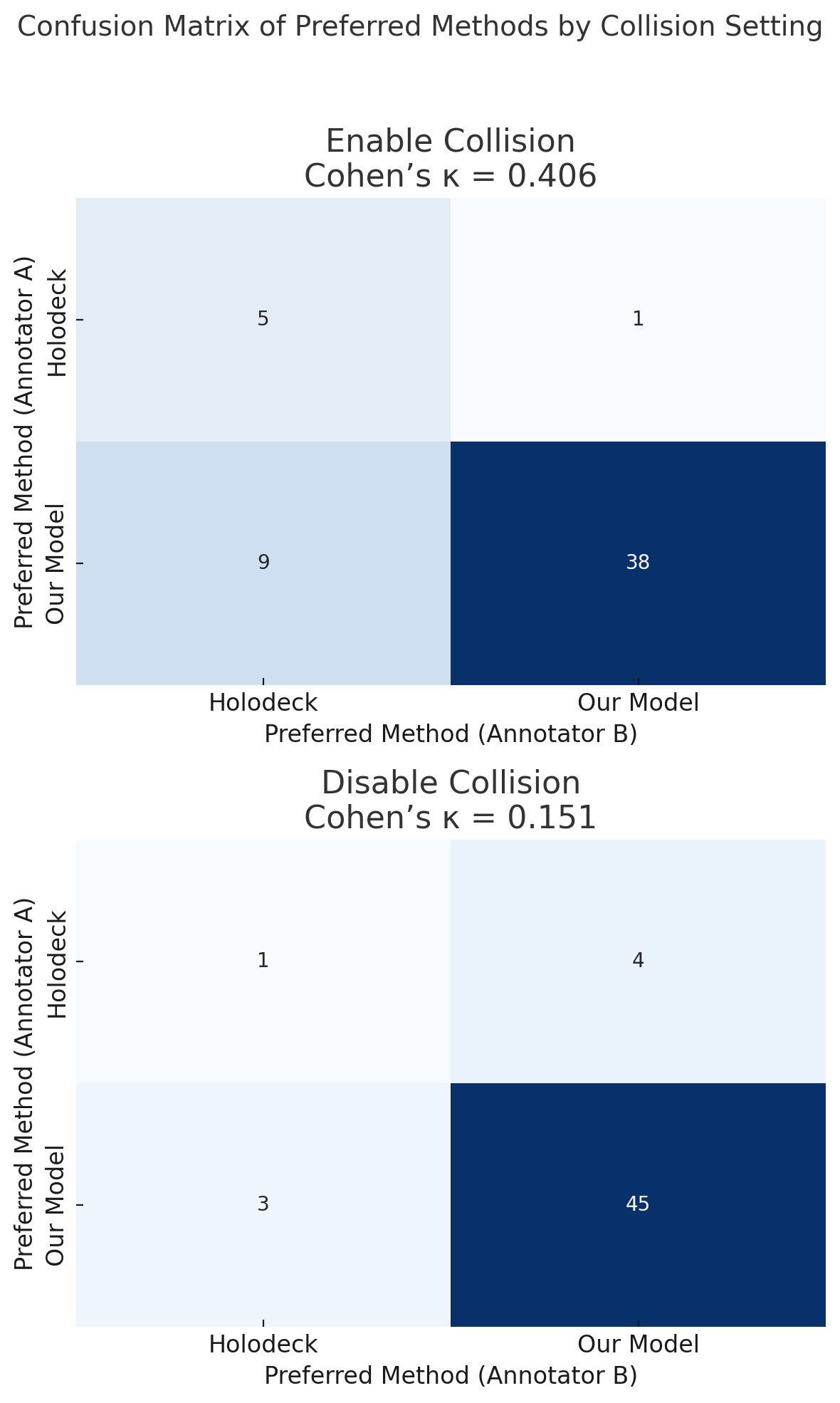}
    \caption{Confusion matrix of binary preference judgments under two collision settings. Each entry indicates how often a model was preferred by two annotators. With collision checking enabled, Cohen’s kappa  0.406 indicates moderate agreement, while disabling collision checking reduces agreement to $\kappa$ = 0.151, suggesting that physical plausibility improves annotator consistency.} 
    \label{fig:task_completion}
    \vspace{1cm}
\end{figure}

Figure~\ref{fig:task_completion} summarizes the results from the binary comparison. Among 53 trials, the collision-aware model was preferred 38 times, while the Holodeck baseline was selected only 6 times. These results suggest that our system significantly improves task satisfaction in human perception.

\subsection{Mean Opinion Score}

To quantify the degree to which the blocking instruction was fulfilled, annotators rated scenes based on three semantic criteria:

\begin{enumerate}
    \item \textbf{Effectiveness:} Are doors effectively blocked by large objects?
    \item \textbf{Arrangement:} Is the object placement intentional and well-structured?
    \item \textbf{Scale Appropriateness:} Are the blocking objects properly sized for their function?
\end{enumerate}

\begin{figure}[t]
    \centering
    \includegraphics[width=\linewidth]{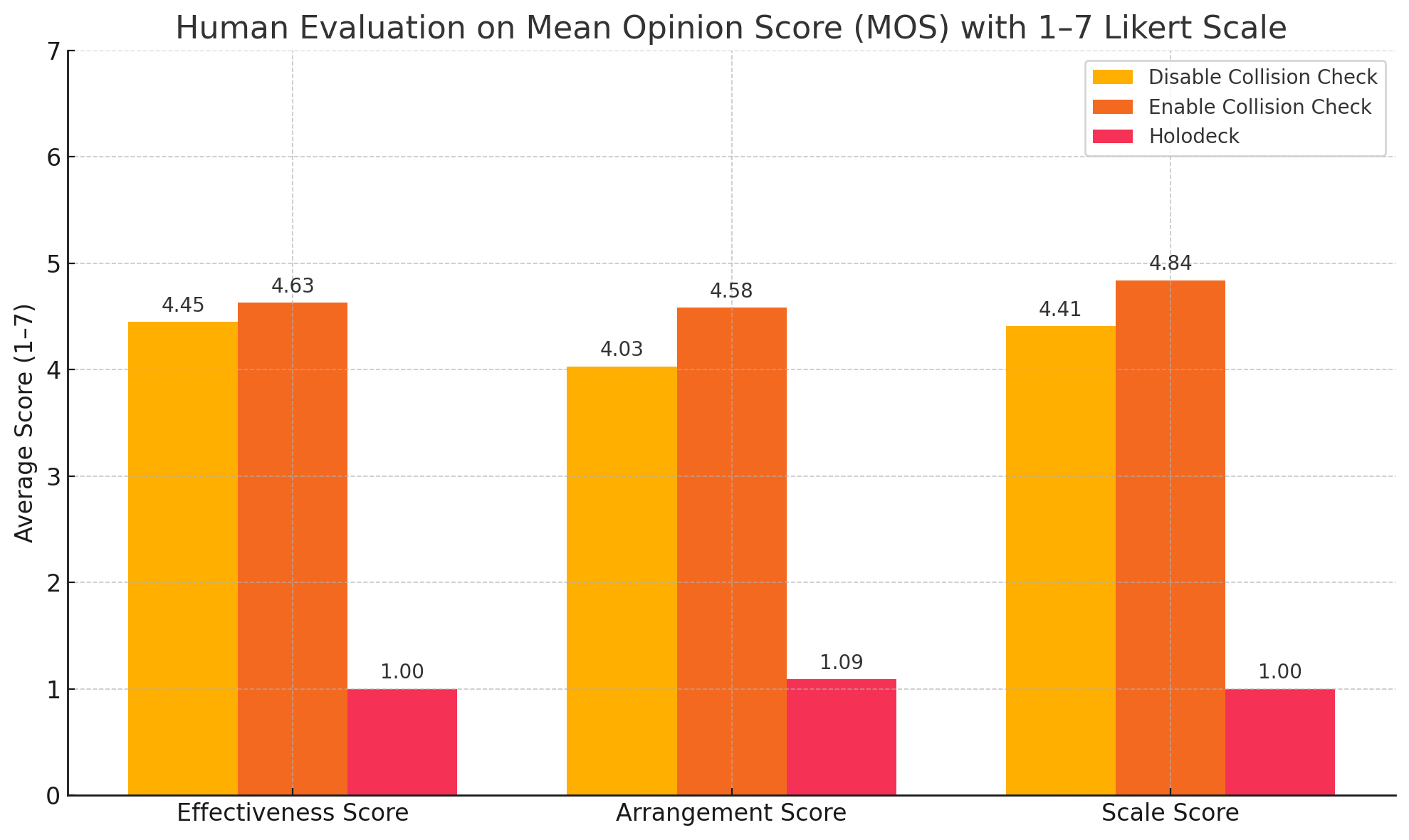}
    \caption{Average Likert scores (1–7 scale) for three goal-oriented evaluation questions. Our collision-aware method outperforms both the collision-disabled version and the Holodeck baseline across all metrics.}
    \label{fig:goal_scores}
  \vspace{1cm}
\end{figure}

As shown in Figure~\ref{fig:goal_scores}, our model with collision checking consistently achieved the highest scores across all dimensions, with average ratings above 4.5 out of 7. In contrast, the Holodeck baseline consistently scored near the minimum, reflecting its lack of semantic alignment.

\subsection{Automatic Evaluation}

To complement human judgments, we conducted an automatic evaluation using two latest LLM models: GPT-4.1 and Gemini 2.5 Pro. Each model was presented with side-by-side renderings of the Holodeck baseline and our edited scenes (both with and without collision checking) and was prompted to select the version that better aligned with the goal prompt. The evaluation setting and scene distribution were kept identical to the human study to ensure direct comparability.

While we initially explored replicating human Likert-scale scoring using language models, we observed that such ratings were often unstable and sensitive to prompt formulation, order effects, and sampling randomness. Consequently, we prioritized Discrete Choice Experiment (DCE), which yielded more consistent and interpretable results.

\begin{figure}[t]
    \centering
    \includegraphics[width=0.7\linewidth]{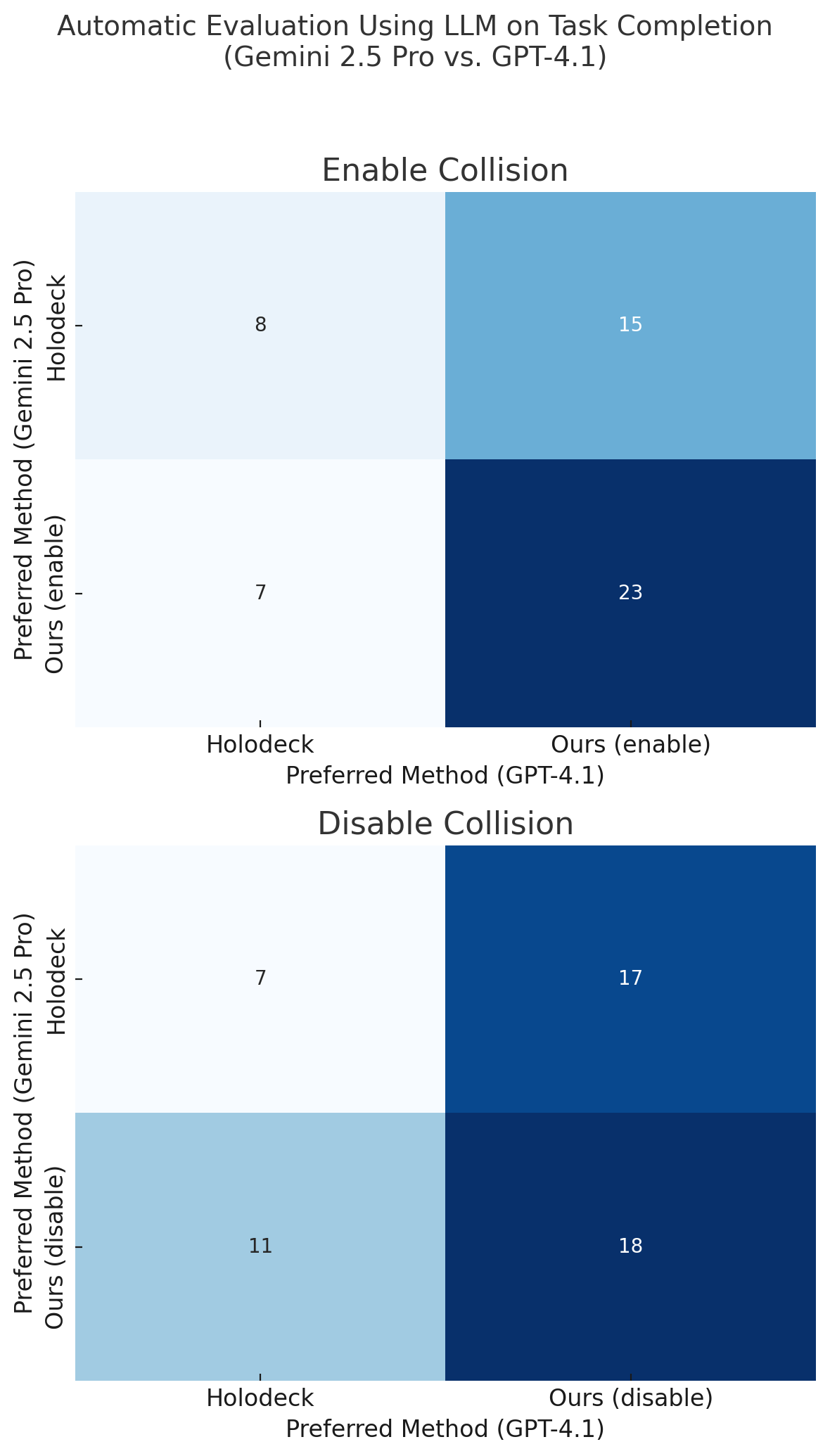}
    \caption{Confusion matrix of LLM judgments under two  collision settings.}
    \label{fig:dce_llm_eval_task_complete}
    \vspace{1cm}
\end{figure}

As summarized in Figure ~\ref{fig:dce_llm_eval_task_complete}, both GPT-4.1 and Gemini strongly preferred our edited scenes over the Holodeck baseline, especially when collision checking was enabled. For instance, GPT-4.1 selected our collision-aware version in 38 out of 53 comparisons. These outcomes reflect a similar trend observed in human preference data, reinforcing the reliability of the edits under both human and machine evaluation.

\begin{figure}[t]
    \centering
    \includegraphics[width=\linewidth]{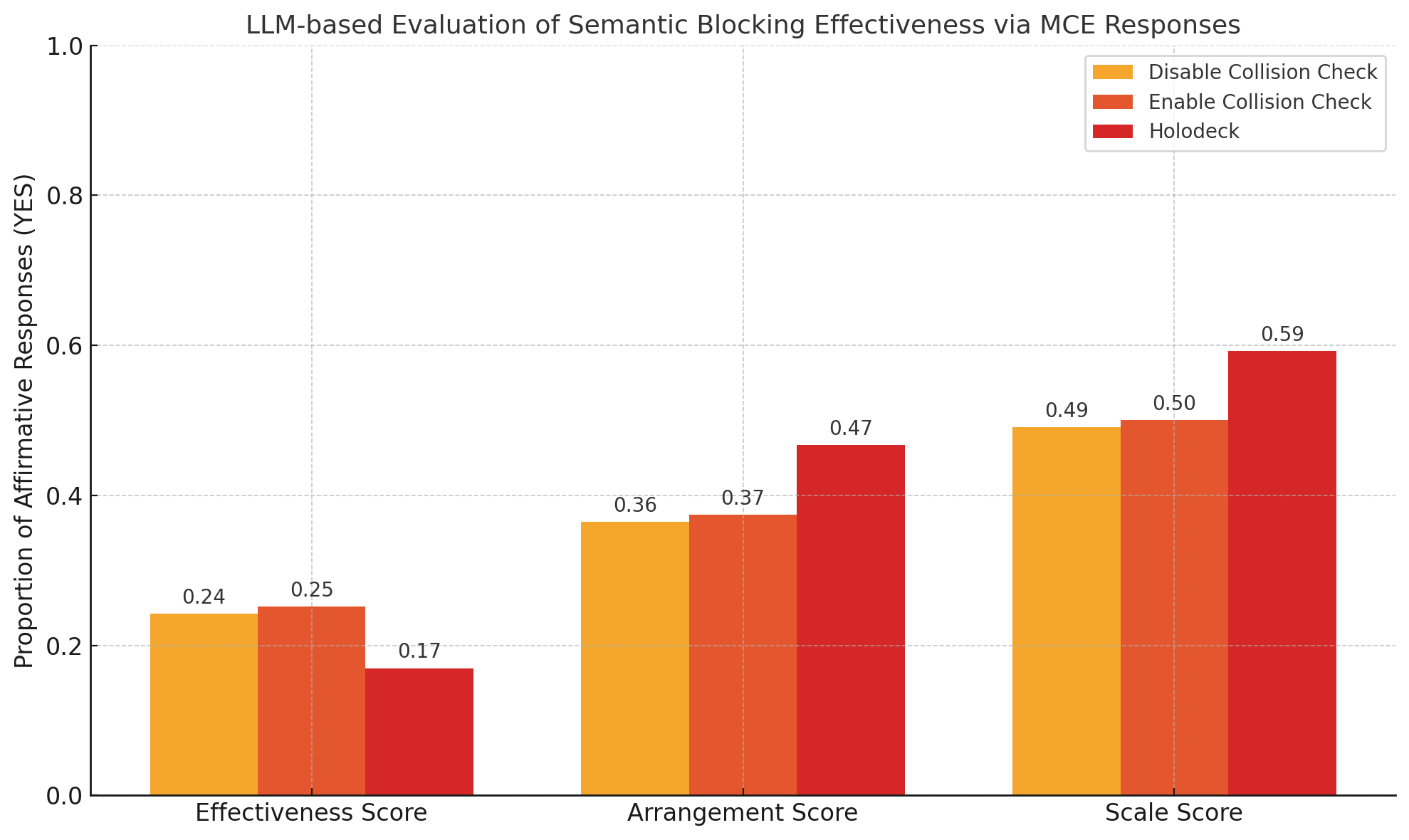}
    \caption{DCE results from GPT-4.1 and Gemini for the three goal-oriented questions.}
    \label{fig:llm_goal_eval}
    \vspace{1cm}
\end{figure}

In addition, we evaluated the semantic quality of the scenes using the same three criteria presented to human annotators: (1) effectiveness in blocking doors, (2) intentionality of object arrangement, and (3) scale appropriateness of blocking objects. As shown in Figure~\ref{fig:llm_goal_eval}, our collision-aware variant achieved the highest score in effectiveness, indicating strong alignment with the intended blocking goal. However, Holodeck outperformed our method in arrangement and scale, which we attribute to its procedurally generated layouts that exhibit higher visual regularity and spatial symmetry, properties that may be implicitly favored by language models. Unlike human annotators, who had access to the goal prompt and could evaluate scenes in context, LLMs may rely more heavily on visual priors learned from web-scale data, leading them to prefer aesthetically balanced scenes over functionally correct ones. This gap highlights the current limitations of LLM-based visual judgment in goal-conditioned settings and reinforces the importance of human-grounded evaluation when assessing safety-critical scene generation.

\subsection{Discussion}
Our evaluation reveals that AgentSGEN meaningfully advances the state of synthetic scene editing by introducing fine-grained semantic control while preserving visual plausibility. Unlike conventional procedural generation pipelines, which often struggle to reconcile realism with goal specificity, our dual-agent architecture, especially when equipped with collision awareness, consistently produces scenes that are not only functionally aligned with safety objectives but also perceived as intentional and well-structured. This indicates that embedding semantic reasoning within the generation loop, rather than relying solely on one-shot prompt-based synthesis, yields more robust and interpretable outcomes for safety-critical scenarios.

Crucially, our approach demonstrates that semantic and perceptual objectives are not inherently at odds. By enabling agents to reason over structured constraints and interactively refine scenes, we maintain high visual fidelity without compromising domain-specific requirements. This integration of symbolic reasoning and reactive control offers a new pathway for scalable, human-aligned synthetic data generation, where safety rules, spatial logic, and perceptual realism are jointly optimized. The strong performance across diverse indoor layouts further underscores the adaptability of our framework and its potential for broader deployment in high-stakes visual AI systems.


\section{Conclusion}

This paper introduces AgentSGEN, a cognitively grounded multi-agent framework that enables semantically controlled, goal-aligned synthetic scene generation for safety-critical applications. By separating high-level reasoning (Evaluator Agent) from low-level execution (Editor Agent), our system achieves fine-grained spatial control while preserving visual plausibility, ann advancement over existing one-shot or purely procedural methods. Through rigorous evaluation, including human assessments across diverse indoor layouts, we demonstrate that AgentSGEN effectively satisfies complex safety constraints without degrading scene realism. The modular design supports scalability and generalization to other domains such as robotics, urban simulation, or healthcare environments.

While our current evaluation is limited by a modest number of annotators, we mitigate potential bias through randomized, blinded sample presentation to ensure fairness and validity. Complementing the human study, we conducted an automatic evaluation using lastests language models (GPT-4.1 and Gemini 2.5 Pro), which independently confirmed the superiority of our approach over baseline methods. The models consistently preferred our edits in binary goal satisfaction tasks, with stronger preference for the collision-aware configuration. The agreement between LLM-based and human evaluations reinforces the reliability of our design choices and suggests that language models can serve as viable auxiliary evaluators for structured scene editing tasks.

Together, these findings position AgentSGEN as a robust, extensible foundation for human-aligned synthetic data generation, capable of meeting nuanced domain constraints while maintaining perceptual and spatial integrity.




\bibliography{mybibfile}

\begin{thebibliography}{27}
\providecommand{\natexlab}[1]{#1}
\providecommand{\url}[1]{\texttt{#1}}
\expandafter\ifx\csname urlstyle\endcsname\relax
  \providecommand{\doi}[1]{doi: #1}\else
  \providecommand{\doi}{doi: \begingroup \urlstyle{rm}\Url}\fi

\bibitem[Bellini-Leite(2022)]{bellini2022dpt}
S.~C. Bellini-Leite.
\newblock Dual process theory: Embodied and predictive; symbolic and classical.
\newblock \emph{Frontiers in Psychology}, 13:\penalty0 805386, 2022.
\newblock \doi{10.3389/fpsyg.2022.805386}.
\newblock URL \url{https://www.frontiersin.org/articles/10.3389/fpsyg.2022.805386/full}.

\bibitem[Blattmann et~al.(2023)Blattmann, Dockhorn, Kulal, Mendelevitch, Kilian, Lorenz, Levi, English, Voleti, Letts, et~al.]{blattmann2023stable}
A.~Blattmann, T.~Dockhorn, S.~Kulal, D.~Mendelevitch, M.~Kilian, D.~Lorenz, Y.~Levi, Z.~English, V.~Voleti, A.~Letts, et~al.
\newblock Stable video diffusion: Scaling latent video diffusion models to large datasets.
\newblock \emph{arXiv preprint arXiv:2311.15127}, 2023.

\bibitem[Bochkovskiy et~al.(2020)Bochkovskiy, Wang, and Liao]{bochkovskiy2020yolov4}
A.~Bochkovskiy, C.-Y. Wang, and H.-Y.~M. Liao.
\newblock Yolov4: Optimal speed and accuracy of object detection.
\newblock \emph{arXiv preprint arXiv:2004.10934}, 2020.

\bibitem[Ding et~al.(2023)Ding, Xu, Arief, Lin, Li, and Zhao]{ding2023survey}
W.~Ding, C.~Xu, M.~Arief, H.~Lin, B.~Li, and D.~Zhao.
\newblock A survey on safety-critical driving scenario generation—a methodological perspective.
\newblock \emph{IEEE Transactions on Intelligent Transportation Systems}, 24\penalty0 (7):\penalty0 6971--6988, 2023.

\bibitem[Esser et~al.(2021)Esser, Rombach, and Ommer]{esser2021taming}
P.~Esser, R.~Rombach, and B.~Ommer.
\newblock Taming transformers for high-resolution image synthesis.
\newblock In \emph{Proceedings of the IEEE/CVF conference on computer vision and pattern recognition}, pages 12873--12883, 2021.

\bibitem[Esser et~al.(2024)Esser, Kulal, Blattmann, Entezari, M{\"u}ller, Saini, Levi, Lorenz, Sauer, Boesel, et~al.]{esser2024scaling}
P.~Esser, S.~Kulal, A.~Blattmann, R.~Entezari, J.~M{\"u}ller, H.~Saini, Y.~Levi, D.~Lorenz, A.~Sauer, F.~Boesel, et~al.
\newblock Scaling rectified flow transformers for high-resolution image synthesis.
\newblock In \emph{Forty-first international conference on machine learning}, 2024.

\bibitem[Hagendorff et~al.(2023)Hagendorff, Fabi, and Kosinski]{hagendorff2023human}
T.~Hagendorff, S.~Fabi, and M.~Kosinski.
\newblock Human-like intuitive behavior and reasoning biases emerged in large language models but disappeared in chatgpt.
\newblock \emph{Nature Computational Science}, 3\penalty0 (10):\penalty0 833--838, 2023.

\bibitem[Hong et~al.(2023)Hong, Zheng, Chen, Cheng, Wang, Zhang, Wang, Yau, Lin, Zhou, et~al.]{hong2023metagpt}
S.~Hong, X.~Zheng, J.~Chen, Y.~Cheng, J.~Wang, C.~Zhang, Z.~Wang, S.~K.~S. Yau, Z.~Lin, L.~Zhou, et~al.
\newblock Metagpt: Meta programming for multi-agent collaborative framework.
\newblock \emph{arXiv preprint arXiv:2308.00352}, 3\penalty0 (4):\penalty0 6, 2023.

\bibitem[Kawar et~al.(2023)Kawar, Zada, Lang, Tov, Chang, Dekel, Mosseri, and Irani]{kawar2023imagic}
B.~Kawar, S.~Zada, O.~Lang, O.~Tov, H.~Chang, T.~Dekel, I.~Mosseri, and M.~Irani.
\newblock Imagic: Text-based real image editing with diffusion models.
\newblock In \emph{Proceedings of the IEEE/CVF conference on computer vision and pattern recognition}, pages 6007--6017, 2023.

\bibitem[Kim and Yi(2024)]{kim2024image}
H.~Kim and J.-S. Yi.
\newblock Image generation of hazardous situations in construction sites using text-to-image generative model for training deep neural networks.
\newblock \emph{Automation in Construction}, 166:\penalty0 105615, 2024.

\bibitem[Lee et~al.(2023)Lee, Jeon, Lee, Park, Kim, and Lee]{lee2023game}
H.~Lee, J.~Jeon, D.~Lee, C.~Park, J.~Kim, and D.~Lee.
\newblock Game engine-driven synthetic data generation for computer vision-based safety monitoring of construction workers.
\newblock \emph{Automation in Construction}, 155:\penalty0 105060, 2023.

\bibitem[Liu et~al.(2023)Liu, Zhang, Li, Liu, and Yang]{liu2023dynamic}
Z.~Liu, Y.~Zhang, P.~Li, Y.~Liu, and D.~Yang.
\newblock Dynamic llm-agent network: An llm-agent collaboration framework with agent team optimization.
\newblock \emph{arXiv preprint arXiv:2310.02170}, 2023.

\bibitem[Mou et~al.(2024)Mou, Wang, Xie, Wu, Zhang, Qi, and Shan]{mou2024t2i}
C.~Mou, X.~Wang, L.~Xie, Y.~Wu, J.~Zhang, Z.~Qi, and Y.~Shan.
\newblock T2i-adapter: Learning adapters to dig out more controllable ability for text-to-image diffusion models.
\newblock In \emph{Proceedings of the AAAI conference on artificial intelligence}, volume~38, pages 4296--4304, 2024.

\bibitem[Neuhausen et~al.(2020)Neuhausen, Herbers, and K{\"o}nig]{neuhausen2020synthetic}
M.~Neuhausen, P.~Herbers, and M.~K{\"o}nig.
\newblock Synthetic data for evaluating the visual tracking of construction workers.
\newblock In \emph{Construction Research Congress 2020}, pages 354--361. American Society of Civil Engineers Reston, VA, 2020.

\bibitem[Podell et~al.(2023)Podell, English, Lacey, Blattmann, Dockhorn, M{\"u}ller, Penna, and Rombach]{podell2023sdxl}
D.~Podell, Z.~English, K.~Lacey, A.~Blattmann, T.~Dockhorn, J.~M{\"u}ller, J.~Penna, and R.~Rombach.
\newblock Sdxl: Improving latent diffusion models for high-resolution image synthesis.
\newblock \emph{arXiv preprint arXiv:2307.01952}, 2023.

\bibitem[Quattoni and Torralba(2009)]{mit_53_scenes_quattoni2009recognizing}
A.~Quattoni and A.~Torralba.
\newblock Recognizing indoor scenes.
\newblock In \emph{2009 IEEE Conference on Computer Vision and Pattern Recognition}, pages 413--420. IEEE, 2009.
\newblock \doi{10.1109/CVPR.2009.5206537}.

\bibitem[Raistrick et~al.(2024)Raistrick, Mei, Kayan, Yan, Zuo, Han, Wen, Parakh, Alexandropoulos, Lipson, et~al.]{raistrick2024infinigen}
A.~Raistrick, L.~Mei, K.~Kayan, D.~Yan, Y.~Zuo, B.~Han, H.~Wen, M.~Parakh, S.~Alexandropoulos, L.~Lipson, et~al.
\newblock Infinigen indoors: Photorealistic indoor scenes using procedural generation.
\newblock In \emph{Proceedings of the IEEE/CVF Conference on Computer Vision and Pattern Recognition}, pages 21783--21794, 2024.

\bibitem[Rombach et~al.(2022)Rombach, Blattmann, Lorenz, Esser, and Ommer]{rombach2022high}
R.~Rombach, A.~Blattmann, D.~Lorenz, P.~Esser, and B.~Ommer.
\newblock High-resolution image synthesis with latent diffusion models.
\newblock In \emph{Proceedings of the IEEE/CVF conference on computer vision and pattern recognition}, pages 10684--10695, 2022.

\bibitem[Sauer et~al.(2024)Sauer, Lorenz, Blattmann, and Rombach]{sauer2024adversarial}
A.~Sauer, D.~Lorenz, A.~Blattmann, and R.~Rombach.
\newblock Adversarial diffusion distillation.
\newblock In \emph{European Conference on Computer Vision}, pages 87--103. Springer, 2024.

\bibitem[Shorten and Khoshgoftaar(2019)]{shorten2019survey}
C.~Shorten and T.~M. Khoshgoftaar.
\newblock A survey on image data augmentation for deep learning.
\newblock \emph{Journal of big data}, 6\penalty0 (1):\penalty0 1--48, 2019.

\bibitem[Song et~al.(2023)Song, He, Li, Ma, Ming, Mao, Pei, Peng, Hu, Yao, et~al.]{song2023synthetic}
Z.~Song, Z.~He, X.~Li, Q.~Ma, R.~Ming, Z.~Mao, H.~Pei, L.~Peng, J.~Hu, D.~Yao, et~al.
\newblock Synthetic datasets for autonomous driving: A survey.
\newblock \emph{IEEE Transactions on Intelligent Vehicles}, 9\penalty0 (1):\penalty0 1847--1864, 2023.

\bibitem[Tran et~al.(2025)Tran, Dao, Nguyen, Pham, O'Sullivan, and Nguyen]{tran2025multi}
K.-T. Tran, D.~Dao, M.-D. Nguyen, Q.-V. Pham, B.~O'Sullivan, and H.~D. Nguyen.
\newblock Multi-agent collaboration mechanisms: A survey of llms.
\newblock \emph{arXiv preprint arXiv:2501.06322}, 2025.

\bibitem[Wang et~al.(2024)Wang, Ma, Feng, Zhang, Yang, Zhang, Chen, Tang, Chen, Lin, et~al.]{wang2024survey}
L.~Wang, C.~Ma, X.~Feng, Z.~Zhang, H.~Yang, J.~Zhang, Z.~Chen, J.~Tang, X.~Chen, Y.~Lin, et~al.
\newblock A survey on large language model based autonomous agents.
\newblock \emph{Frontiers of Computer Science}, 18\penalty0 (6):\penalty0 186345, 2024.

\bibitem[Yang et~al.(2024)Yang, Sun, Weihs, VanderBilt, Herrasti, Han, Wu, Haber, Krishna, Liu, et~al.]{yang2024holodeck}
Y.~Yang, F.-Y. Sun, L.~Weihs, E.~VanderBilt, A.~Herrasti, W.~Han, J.~Wu, N.~Haber, R.~Krishna, L.~Liu, et~al.
\newblock Holodeck: Language guided generation of 3d embodied ai environments.
\newblock In \emph{Proceedings of the IEEE/CVF Conference on Computer Vision and Pattern Recognition}, pages 16227--16237, 2024.

\bibitem[Zhang et~al.(2017)Zhang, Cisse, Dauphin, and Lopez-Paz]{zhang2017mixup}
H.~Zhang, M.~Cisse, Y.~N. Dauphin, and D.~Lopez-Paz.
\newblock mixup: Beyond empirical risk minimization.
\newblock \emph{arXiv preprint arXiv:1710.09412}, 2017.

\bibitem[Zhang et~al.(2023{\natexlab{a}})Zhang, Rao, and Agrawala]{zhang2023adding}
L.~Zhang, A.~Rao, and M.~Agrawala.
\newblock Adding conditional control to text-to-image diffusion models.
\newblock In \emph{Proceedings of the IEEE/CVF international conference on computer vision}, pages 3836--3847, 2023{\natexlab{a}}.

\bibitem[Zhang et~al.(2023{\natexlab{b}})Zhang, Han, Ghosh, Metaxas, and Ren]{zhang2023sine}
Z.~Zhang, L.~Han, A.~Ghosh, D.~N. Metaxas, and J.~Ren.
\newblock Sine: Single image editing with text-to-image diffusion models.
\newblock In \emph{Proceedings of the IEEE/CVF Conference on Computer Vision and Pattern Recognition}, pages 6027--6037, 2023{\natexlab{b}}.

\end{thebibliography}

\end{document}